\documentclass{article}

\usepackage[round]{natbib}

\usepackage[utf8]{inputenc} 
\usepackage[T1]{fontenc}    
\usepackage{hyperref}       
\usepackage{url}            
\usepackage{booktabs}       
\usepackage{amsfonts}       
\usepackage{nicefrac}       
\usepackage{microtype}      
\usepackage{amsmath}
\usepackage{amssymb}
\usepackage{graphicx} 
\usepackage{float}
\usepackage{enumitem}
\usepackage{lipsum} 
\usepackage{algorithm}
\usepackage[noend]{algpseudocode}
\usepackage{subfig} 
\usepackage{color}
\usepackage{bbm}

\newcommand{\remove}[1]{}
\usepackage{tikz}
\usepackage[toc,page]{appendix}
\usepackage{xfrac}
\usepackage{float}
\usepackage[a4paper, total={6in, 8in}]{geometry}

\definecolor{right}{HTML}{697c2a}
\definecolor{down}{HTML}{000000}

\title{Independently Controllable Factors}

\makeatletter

\renewcommand*{\@fnsymbol}[1]{\ifcase#1\or*\else\@arabic{\numexpr#1-1\relax}\fi}
\makeatother

\author{
Valentin Thomas\,\thanks{Equal contribution, random order}\,\ \thanks{MILA, Universit\'{e} de Montr\'{e}al}\,\ \thanks{ElementAI}
\and
Jules Pondard\,\footnotemark[1]\,\ \footnotemark[2]\,\ \footnotemark[3]\,\ \thanks{ENS Paris}
\and
Emmanuel Bengio\,\footnotemark[1]\,\ \thanks{McGill University}
\and 
Marc Sarfati\footnotemark[2]\,\ \thanks{\'{E}cole Polytechnique}
\and
Philippe Beaudoin\footnotemark[3]
\and
Marie-Jean Meurs\thanks{Universit\'{e} du Qu\'{e}bec \`{a} Montr\'{e}al}
\and
Joelle Pineau\footnotemark[5] 
\and
Doina Precup\footnotemark[5]
\and
Yoshua Bengio\footnotemark[2]\,\ \thanks{CIFAR Senior Fellow}
}

\newif\ifnomulti
\nomultitrue

\newif\ifnotransfer
\notransfertrue

\begin{document}

\maketitle

\begin{abstract}
  It has been postulated that a good representation is one that disentangles the underlying explanatory factors of variation. However, it remains an open question what kind of training framework could potentially achieve that. Whereas most previous work focuses on the static setting (e.g., with images), we postulate that some of the causal factors could be discovered if the learner is allowed to interact with its environment. The agent can experiment with different actions and observe their effects. More specifically, we hypothesize that some of these factors correspond to aspects of the environment which are independently controllable, i.e., that there exists a policy and a learnable feature for each such aspect of the environment, such that this policy can yield changes in that feature with minimal changes to other features that explain the statistical variations in the observed data. We propose a specific objective function to find such factors and verify experimentally that it can indeed disentangle independently controllable aspects of the environment without any extrinsic reward signal.
\end{abstract}

\section{Introduction}
Whether in static or dynamic environments, decision making for real world problems is often confronted with the hard challenge of finding a ``good'' representation of the problem. 
In the context of supervised or semi-supervised learning, it has been argued~\citep{Bengio-2009-book} that good representations separate out underlying explanatory factors, which may be causes of the observed data.
In such problems, feature learning often involves mechanisms such as autoencoders \citep{hinton2006reducing,VincentICML2008}, which find latent features that explain the observed data. In interactive environments, the temporal dependency between successive observations creates a new opportunity to notice causal structure in data which may not be apparent using only observational studies. The need to experiment in order to discover causal relationships has already been well explored in psychology (e.g.~\cite{gopnik}). In reinforcement learning, several approaches explore  mechanisms that push the internal representations of  learned models to be ``good'' in the sense that they provide better control (see \S\ref{sec:related}), and control is a particularly important causal relationship between an agent and elements of its environment.

We propose and explore a more direct mechanism for representation learning, which explicitly links an agent's control over its environment with its internal feature representations. Specifically, we hypothesize that some of the factors explaining  variations in the data correspond to aspects of the world that can be controlled by the agent. For example, an object that could be pushed around or picked up independently of others is an independently controllable aspect of the environment. Our approach therefore aims to jointly discover a set of features (functions of the environment state) and policies (which change the state) such that each policy controls the associated feature while leaving the other features unchanged as much as possible. 
In \S\ref{sec:icf} and \S \ref{sec:exp-results} we explain this mechanism and show experimental results for the simplest instantiation of this new principle. In \S\ref{sec:general} we discuss how this principle could be applied more generally, and what are the research challenges that emerge.

\section{Independently controllable features}
\label{sec:icf}

To make the above intuitions concrete, assume that there are factors of variation underlying the  observations coming from an interactive environment that are \emph{independently controllable}. That is, a controllable factor of variation is one for which there exists a policy which will modify that factor only, and not the others. 
For example, the object associated with a set of pixels could be acted on independently from other objects, which would explain variations in its pose and scale when we move it around while leaving the others generally unchanged. The object position in this case is a \emph{factor of variation}.
What poses a challenge for discovering and mapping such factors into computed features is the fact that the factors are not explicitly observed. Our goal is for the agent to autonomously discover such factors -- which we call~\textbf{independently controllable features} -- along with policies that control them.
While these may seem like strong assumptions about the nature of the environment, we argue that these assumptions are similar to regularizers, and are meant to make a difficult learning problem (that of learning good representations which disentangle underlying factors) better constrained.

There are many possible ways to express the preference for learning independently controllable features as an objective. \S \ref{sec:pol-sel} proposes such an objective for a simple scenario. \S \ref{sec:exp-emmanuel} illustrates the effect of this objective when all the features of the environment are simple and controllable by the agent.
Moreover, in \S \ref{sec:sel-only}, we show that by itself the objective we propose is strong enough to recover underlying factors of variation without additional reconstruction loss.
In \S \ref{sec:polemb}, we aim to generalize such an objective for a continuous representation of factors and policies. In \S \ref{sec:exp-mb}, we present our experiments on the Mazebase domain \citep{sukhbaatar2015mazebase}. \S \ref{sec:exp-mb-cpe} shows that, using these continuous embeddings, we are able to disentangle the latent space and the controllable factors. In \S \ref{sec:exp-mb-ppi}, we show how the learnt representations can be used for planning and for inferring the sequence of actions performed between two states.

\subsection{Capturing the main factors of variation}
Since not all factors of variation present in the data are controllable, we propose to combine two objectives: (1) one to encourage the learned representation to capture the main factors of variation, and (2) one to encourage the representation to be structured so that the controllable factors are disentangled from each other and from other factors. Any common method for representation learning could be used for (1); for simplicity we use a simple autoencoder framework throughout this paper \citep{hinton2006reducing}.
The encoder and decoder of the autoencoder are viewed as function approximators $f,g$ with parameters $\theta$ such that $f:X\to H$ maps the input space to some \textit{latent space} $H \subseteq \mathbb{R}^n$, and $g:H\to X$ maps back to the input space $X \subset \mathbb{R}^d$. Autoencoders are trained to minimize the discrepancy between $x$ and $g(f(x))$, a.k.a. the reconstruction error, e.g.,:
\begin{equation*}
    \min_\theta \tfrac{1}{2} \|x-g(f(x))\|_2^2
\end{equation*}
We call $f(x)=h\in H \subset \mathbb{R}^n$ the latent feature representation of $x$, with $n$ features.

It is common in the case of a vanilla autoencoder to assume that  $n \ll d$. This causes $f$ and $g$ to perform dimensionality reduction of $X$, i.e. \textit{compression}, since there is a dimension bottleneck through which information about the input data must pass. Often, this bottleneck forces the optimization procedure to uncover principal factors of variation of the data on which they are trained. However, this does not necessarily imply that the different
components of the vector $h=f(x)$ are individually meaningful. In fact, note that for any bijective transformation $T$, we
could obtain the same reconstruction error by
replacing $f$ by $T \circ f$ and $g$ by $g \circ T^{-1}$, so we should not expect any form of disentangling
of the factors of variation unless some additional
constraints or penalties are imposed on $h$. This
motivates the approach we are about to present. Specifically, we have a preference for
policies that can separately influence one of the coordinates of $h$, and we want to express a preference for learning representations that make such policies possible.

Note that there may be several other ways to discover and disentangle underlying factors of variation. Many deep generative models, including variational autoencoders~\citep{Kingma+Welling-ICLR2014} and other descendants of the Helmholtz machine~\citep{Dayan-et-al-1995}, generative adversarial
networks~\citep{Goodfellow-et-al-NIPS2014} or non-linear versions of ICA~\citep{Dinh-et-al-2014,Hyvarinen+Morioka-NIPS2016} attempt to disentangle the underlying factors of variation by assuming that their joint distribution (marginalizing out the observed $x$) factorizes, i.e., that they are marginally independent. Here we explore another direction, trying to exploit the ability of a learning agent to act in the world in order impose a further constraint on the representation.

\subsection{Disentangling independently controllable factors in the simplest case}
\label{sec:pol-sel}

Consider the following simple scenario: we train an autoencoder $f,g$ producing $K$  latent features  $f_k:X\to \mathbb{R},\;k\in[K]$. In tandem with these features we train $K$ policies, denoted $\pi_k(a|s)$, that map an agent's observation $s$ to a categorical distribution over a set of actions $a$. Autoencoders can learn relatively arbitrary feature representations, but we would like many of
these features to correspond to controllable
factors in the learner's environment. Specifically, we would like policy $\pi_k$ to cause a change only in $f_k$ and not in any other features. We think of $f_k$ and $\pi_k$ as a feature-policy pair.

In order to quantify the change in $f_k$ when actions are taken according to $\pi_k$, we define the  \emph{selectivity} of a feature as:
\begin{equation}
    sel(s,a,k) = \mathbb{E}_{s'\sim \mathcal{P}_{ss'}^a} \left[ \frac{|f_k(s')-f_k(s)|}{\sum_{k'}|f_{k'}(s')-f_{k'}(s)|} \right]. \label{eq:sel}
\end{equation}
where $s$,$s'$ are successive raw state representations (e.g. pixels), $a$ is the action, and $\mathcal{P}_{ss'}^a$ is the environment transition distribution from $s$ to $s'$ under action $a$. The normalization factor in the denominator of the above equation ensures that the selectivity of $f_k$ is maximal when \textit{only that single feature} $f_k$ changes as a result of some action. 

By having an objective that maximizes selectivity \textit{and} minimizes the autoencoder objective, we can ensure that the features learned can both capture the main factors of variation in the data and recover independently controllable factors. Hence, we define the following objective, which can be minimized jointly on $\pi_k$, $f$ and $g$, via stochastic gradient descent:

\begin{equation}
     \underbrace{\mathbb{E}_s [\tfrac{1}{2} || s - g(f(s))||^2_2]}_{\mathcal{L}_{ae}\text{ the reconstruction error}}\ -\ \lambda \underbrace{\sum_k \mathbb{E}_{s}[\sum_a \pi_k(a|s) sel(s, a, k)   ].}_{\mathcal{L}_{sel} \text{ the disentanglement objective}}  \label{eq:full-objective}
\end{equation}
Here one can think of $sel(s,a,k)$ as the reward signal $R_k(s,a)$ of a control problem, and the expected reward $\mathbb{E}_{a\sim \pi_k}[R_k]$ is maximized by finding the optimal set of policies $\pi_k$.

Note that many variations of this objective are possible. For example it is also possible to have \textit{directed} selectivity: by using $\max\{0,f_k'-f_k\}$ (denoted $|f_k'-f_k|_+$) or simply  $f_k'-f_k$ instead of the absolute value $|f_k'-f_k|$ in the numerator of \eqref{eq:sel}, the policies must learn to increase the learned latent feature rather than simply change it. This may be useful if the policy to gradually increase a feature is distinct from the policy that decreases it. Using log-selectivity, $\log sel$, or this sharpened form, $\log (sel/(1-sel))$, may also lead to easier optimization.

The learning algorithm we propose is summarized in Algorithm \ref{alg1}, where $W_f$ and $W_g$ are the parameters of $f,g$ and $\theta_k$ the parameters of $\pi_k$.

\begin{algorithm}                      
\caption{Training an autoencoder with disentangled factors}          
\label{alg1}                           
\begin{algorithmic}[1]                   
    \For{$t=1..T$}
        \State{Sample $s$ from the environment}
        \State{$W_f \gets W_f - \eta_{f} \nabla_{W_f} [ \tfrac{1}{2} ||s -g (f(s))||^2_2]$} 
        \State{$W_g \gets W_g -  \eta_{g} \nabla_{W_g} [ \tfrac{1}{2} ||s -g (f(s))||^2_2]$} 
        \For{$k=1..n$}
            \State{$W_f \gets W_f + \eta_{f} \ \lambda\  \nabla_{W_f} \mathbb{E}_{a \sim \pi_k(\cdot | s)} [  sel(s,a,k) ]$} 
            \State{$\theta_k \gets \theta_k + \eta_{k} \ \lambda \ \nabla_{\theta_k}  \mathbb{E}_{a \sim \pi_k(\cdot | s)} [ sel(s,a,k) ]$}
        \EndFor
    \EndFor
\end{algorithmic}
\end{algorithm}

The gradients on lines 3 and 4 are computed exactly via backpropagation. In our experiments, the gradient on line 6 is also computed by backpropagation and sampling of the expectation, while the gradient on line 7 is computed 
with the REINFORCE \citep{glynn1987likelilood,williams1992simple} estimator: 
\begin{equation*} 
   \nabla_{\theta_k} \mathbb{E}_{a \sim \pi_k(\cdot | s)}[ sel(s, a, k) ] = \mathbb{E}_{a \sim \pi_k(\cdot | s)}[  (sel(s, a, k) - b(s)) \cdot \nabla_{\theta_k} \log \pi_k(a|s)],
   \end{equation*}
where $b(s)$ is a baseline function, which can for example be chosen to be the mean reward or an estimate of the value of the state.

\subsection{From enumerated factors to continuous embeddings of individual factors}
\label{sec:polemb}
A limitation of the approach in Algorithm~\ref{alg1} is that it requires the set of potentially controllable factors
to be small and enumerated. This makes sense in a simple environment where we always have the same set of
objects in the scene. But in more realistic environments, the number of possible objects present in the
set can be combinatorially large (and better described by notions such as types), 
while an individual scene will only comprise a finite number of {\em instances} of such objects.
Therefore, instead of indexing the possible factors by an integer, we propose to index
them by an embedding, i.e., a real-valued vector. 
In the last section, we enforced variations in the environment to be captured by a coordinate of $h = f(s)$. We can view this as having a set of $k$ attribute variations $A(h' - h, k) = |h' - h|_k$ \remove{$ = \mathbbm{1}_k^\top h = h_k$ \footnote{where $\mathbbm{1}_k$ is a column one-hot vector whose $k$-th coordinate is $1$.} }who are influenced separately by the policies $\pi_k$. We now relax this assumption by indexing this set by a learned real-valued vector $\phi$ leading to a continuous set of attributes $A(h' - h,\phi) \in \mathbb{R}$.
The idea of mapping symbolic entities to
a distributed representation is one of the key ingredients of the success of deep learning \citep{Goodfellow-et-al-2016},
and can be exploited here as well.

\paragraph{Selecting attributes}
Conditioned on a scene representation $h$, a distribution of policies are feasible. Samples from this distribution represent ways to modify the scene and thus may trigger an internal selectivity reward signal. For instance, $h$ might represent a room with objects such as a light switch. $\phi = \phi(h,z)$ can be thought of as the distributed representation for the ``name'' of an underlying factor, to which is associated a policy and a value.  In this setting, the light in a room could be a factor that could be either on or off. It could be associated with a policy to turn it on, and a binary value referring to its state, called an attribute or a feature value.
We wish to jointly learn the policy $\pi_\phi(\cdot|s)$ that modifies the scene, so as to control the corresponding value of the attribute in the scene, whose variation is computed by an attribute variation selector function $A(h' - h, \phi) \in \mathbb{R}$. In order to get a distribution of such embeddings, we compute $\phi(h,z)$ as a function of $h$ and some random noise $z$.

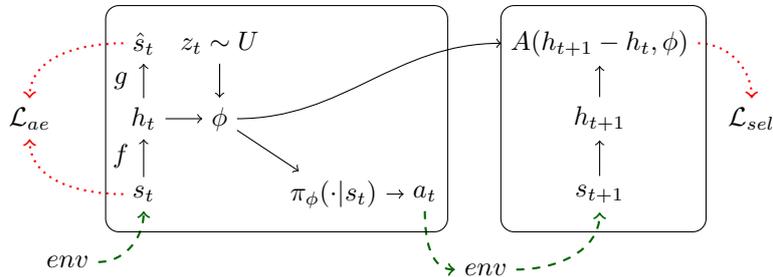
\begin{figure}[!ht]
\centering
\begin{tikzpicture}

\def \pA {0}
\def \pB {6}
\def \pmid {2.5}

\node[] (st) at (\pA,0) {$s_t$};
\node[] (ht) at (\pA,1) {$h_t$};
\node (st_rec) at (\pA,2) {$\hat{s}_t$};

\draw[->,draw=black] (st) to (ht);
\draw[->,draw=black] (ht) to (st_rec);
\node at (-.3,1.5) {$g$};
\node at (-.3,0.5) {$f$};

\node (env) at (-1,-0.9) {$env$};
\draw[->,draw=black!60!green,dashed,thick] (env) to[out=0,in=-90] (st);

\node[] (st1) at (\pB,0) {$s_{t+1}$};
\node[] (ht1) at (\pB,1) {$h_{t+1}$};
\draw[->,draw=black] (st1) to (ht1);

\node[] (pi0) at (\pmid,0) {$\pi_\phi(\cdot|s_t)$};
\node[] (phi0) at (\pmid-1.5,1) {$\phi$};
\node[] (z0) at (\pmid-1.5,2) {$z_t \sim U$};
\node[] (A1) at (\pB,2) {$A(h_{t+1}-h_t,\phi)$};
\draw[->,draw=black] (z0) to (phi0);
\draw[->,draw=black] (ht) to (phi0);
\draw[->,draw=black] (phi0) to (pi0);
\draw[->,draw=black] (phi0) to[in=180,out=0] (A1);
\draw[->,draw=black] (ht1) to (A1);

\begin{scope}[]

\end{scope}
\node[] (aT) at (\pmid+1.2,-0) {$a_t$};
\node[] (env1) at (\pmid+2,-1) {$env$};
\draw[->] (pi0) to (aT);
\draw[->,draw=black!60!green,dashed,thick] (aT) to[out=-90,in=180] (env1);
\draw[->,draw=black!60!green,dashed,thick] (env1) to[out=0,in=-90] (st1);

\node[] (lossAE) at (\pA-1.5,1) {$\mathcal{L}_{ae}$};
\draw[->,draw=red,dotted,thick] (st) to[out=180,in=-90] (lossAE);
\draw[->,draw=red,dotted,thick] (st_rec) to[out=180,in=90] (lossAE);

\node[] (lossSel) at (\pB+2,1) {$\mathcal{L}_{sel}$};
\draw[->,draw=red,dotted,thick] (A1) to[out=0,in=90] (lossSel);

\draw
  {[rounded corners=5pt](-0.5,-0.5)  --
  (\pmid+1.5,-0.5)  -- 
  (\pmid+1.5,2.5) --
  (-0.5,2.5) --
  cycle};
\draw
  {[rounded corners=5pt](-1.3+\pB,-0.5)  --
  (1.4+\pB,-0.5)  -- 
  (1.4+\pB,2.5) --
  (-1.3+\pB,2.5) --
  cycle};
\end{tikzpicture}
\caption{The proposed distributed representation architecture. $\mathcal{L}_{ae}$ and $\mathcal{L}_{sel}$ are the reconstruction and selectivity objectives respectively.}
\label{fig:polemb_model}
\end{figure}

In this scenario, one strategy to determine whether some selected attribute variation $A(h' - h,\phi)$ evolves \emph{independently} from other attributes variations is to compare its value (in expectation over the policy actions) to the values obtained with other $\phi'$ factors. We thus compute the following selectivity that acts as an intrinsic reward signal, generalizing \eqref{eq:sel}:

\begin{equation}
    sel(h, \phi) = \mathbb{E}_{a \sim \pi_\phi(\cdot | s), s' \sim \mathcal{P}_{s s'}^a} \left[ \frac{A(h' - h,\phi)}{\mathbb{E}_{\phi'} | A(h' - h,\phi')|} \right], \label{eq:sel_cont}
\end{equation}

where $h'=f(s')$. We approximate the expectation over $\phi'$ by sampling a fixed number of factor embeddings. This model is then trained by jointly minimizing the autoencoder reconstruction cost $\mathcal{L}_{ae}$ and the disentanglement objective $\mathcal{L}_{sel}$ as depicted in Figure~\ref{fig:polemb_model}.



\paragraph{Implementing an attribute selector}
Ideally $A(h' - h,\phi)$ could be an arbitrary function, e.g. a neural network, but such function may be harder to optimize. Instead, we observe that in the discrete case mentioned previously, using $A(h' - h,\phi)$ to select attribute $k$ is equivalent to $\phi^\top |h' - h|$ where $\phi$ is a one-hot vector at index $k$. 
One simple step towards continuous embeddings is to relax this constraint, and let $\phi$ be a function of $h$ and random vector $z$, drawn from uniform distribution, and compute $A$ as $A(h' - h,\phi) = \phi(h,z)^\top |h' - h|$. However, in most of our experiments, we used a gaussian kernel: 
$A(h' - h, \phi) = \mbox{exp}({-||h' - h - \phi||^2/(2 \sigma^2)})$ because of the better numerical stability it provides. 


Unlike in the finite case, we are not sampling uniformly over policies $\pi_k$, as we now let a neural network choose $\phi$'s probability distribution. This could lead to exploration issues. We demonstrate that simple strategies allow for a network to learn simple distributions in the experiments of \S \ref{sec:exp-mb}.









\ifnomulti
\remove{
\fi
\subsection{Multi-step objectives for selectivity}
Does not work for now, selectivity is not converging nor saturating.

Another challenge in control is planning for the future, in this case planning for selectivity. There are many possible approaches that can be taken to have a proper multi-step objective.

For each policy $\pi_k$, our complete return on a trajectory is the discounted sum of the rewards $G^k_T = \sum \limits_{t \le T} \gamma^t r^k_t$.

\begin{equation}
    r_t(a_t, k) = sel_k(s_t, a_t) = \mathbb{E}_{a_t \sim \pi_k,\; s_{t+1} \sim p(\cdot | s_t, a_t}) \left[\frac{|f_k(s_{t+1})-f_k(s_t)|_+}{\sum_{k'}|f_{k'}(s_{t+1})-f_{k'}(s_t)|} \right]
\end{equation}
where $t$ is the current time step.



where $s_T$ is a ``terminal'' state. The agent may have to have an extra action where it chooses to be in such ``terminal'' state, in the sense that the agent recognizes it can no longer keep modifying feature $f_k$.

Another view would be to force the agent to ignore intermediate values of selectivity (i.e. it ignores what other attributes were modified during the trajectory), so that only the selectivity between the first and last visited states matters.
Our return is then 
\begin{equation}
      G^k_T = sel(s_0,s_T,a,k) = \mathbb{E}_{\tau = (s_0,...,s_T)\sim \pi_k} \left[\frac{|f_k(s_{T})-f_k(s_0)|_+}{\sum_{k'}|f_{k'}(s_T)-f_{k'}(s_0)|} \right]
\end{equation}
\ifnomulti
}
\fi

\section{Experimental results}
\label{sec:exp-results}
In order to validate that our method learns independently controllable features, we perform several experiments. First, in the most basic gridworld-like setting, an agent is allowed to move around in four directions. This basic domain allows us to verify whether in the discrete case, the learning process disentangles the underlying features and recovers the ground truth properties of the environment.

Then, we show results of our continuous factors embeddings method applied to MazeBase~\citep{sukhbaatar2015mazebase}, as well as how we can use the learned representations to tackle policy inference or planning problems.


\subsection{A simple gridworld}
\label{sec:exp-emmanuel}

\begin{figure}[H]
\centering

\subfloat[]{\includegraphics[width=0.5\columnwidth,trim={0 30px 0 0},clip]{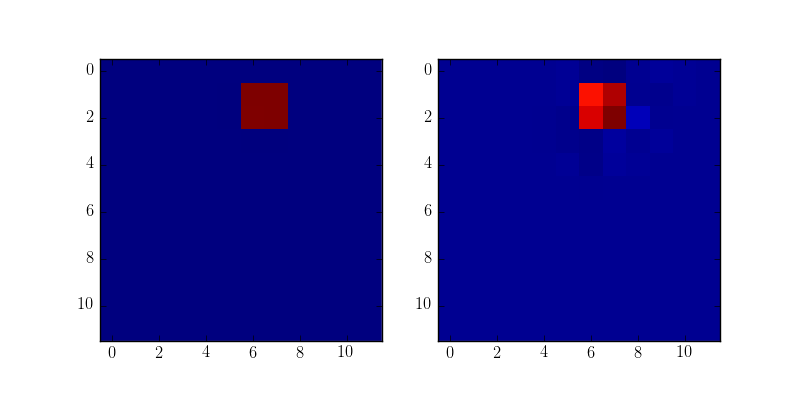}}
\subfloat[]{\includegraphics[width=0.2\columnwidth,trim={75px 0 0 0},clip]{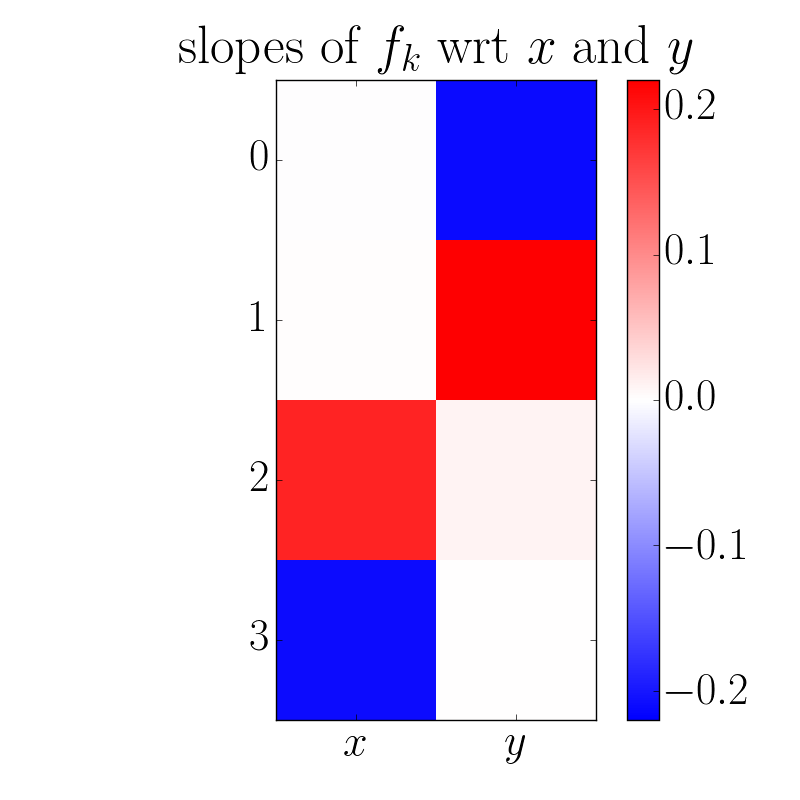}}
\subfloat[]{\includegraphics[width=0.23\columnwidth]{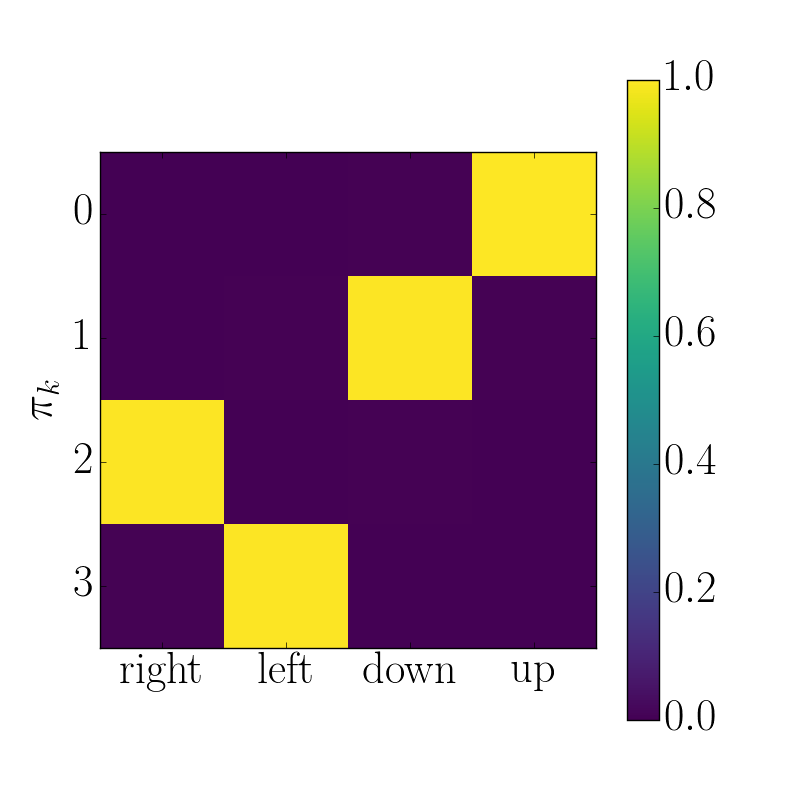}}
\caption{A simple gridworld with 4 actions that push a square left, right, up or down. (a) left is an example ground truth, right is the reconstruction of the model trained with selectivity. (b) The slope of a linear regression of the true features (the real $x$ and  $y$ position of the agent) as a function of each latent feature. White is no correlation, blue and red indicate strong negative or positive slopes respectively. $f_0$ and $f_1$ recover $y$ and $f_2$ and $f_3$ recover $x$. (c) 
Each row is a policy $\pi_k$, each column corresponds to an action (left/right/up/down). Cell $(k,i)$ is the average over $s$ of $\pi_k(a_i|s)$;   
}
\label{fig:squares-env-result}
\end{figure}

Our first experiment is performed on a gridworld-like setting, illustrated in 
Figure \ref{fig:squares-env-result}(a): the agent sees a $2\times 2$ square on a $12\times 12$ pixel grid, and has 4 actions that move it up, down, left or right. By interacting with the environment, an autoencoder 
\footnote{We use the following architecture: $f$ has two $16\times 3 \times 3$ ReLU convolutional layers with stride 2, followed by a fully connected ReLU layer of 32 units, and a $\tanh$ layer of $n=4$ features; $g$ is the transpose architecture of $f$; $\pi_k$ is a softmax policy over 4 actions, computed from the output of the ReLU fully connected layer. We use Adam \citep{kingma2014adam} to perform gradient descent.} 
with directed selectivity (objective \eqref{eq:sel} without absolute value in the numerator) learns latent features that map to the $(x,y)$ position of the square (see Figure \ref{fig:squares-env-result}(b,c)), without ever having explicit access to these values, and while reconstructing its input properly.
In contrast, a plain autoencoder also reconstructs properly but without learning the two latent $(x,y)$ features explicitly. 


Note that in this setting, the learning process is robust to a  stochastic version of the environment -- where with probability $p$ either no action is taken ($s=s'$) or a random action is taken. We have successfully trained models recovering $\pm x$ and $\pm y$ with up to $p=0.5$, using the same architecture but a smaller learning rate.

\subsection{Selectivity as an only objective}
\label{sec:sel-only}
We also find experimentally that training discrete independently controllable features without training the autoencoder objective correctly recovers ground truth features and their associated control policies. Albeit slower than when jointly training an autoencoder, this shows that the objective we propose is strong enough to provide a learning signal for discovering a disentangled latent representation.

\begin{figure}[H]
\centering
\includegraphics[width=\linewidth]{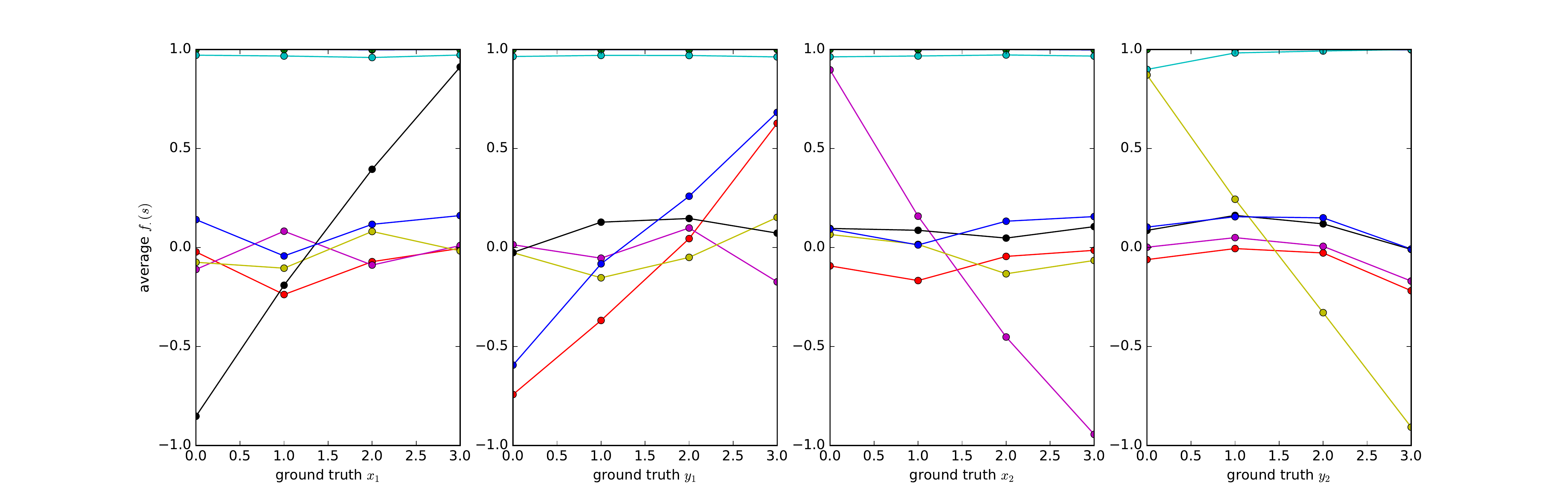}
\caption{
In a gridworld environment with 2 objects (in this case 2 MNIST digits), we know there are 4 underlying features, the $(x_i,y_i)$ position of each digit $i$. Here each of the four plots represents the evolution of the $f_k$'s as a function of their underlying feature, from left to right $x_1$, $y_1$, $x_2$, $y_2$. We see that for each of them, at least one $f_k$ recovers it almost linearly, from the raw pixels only.
}
\label{fig:gridworld-only-sel}
\end{figure}

We train such a model on a gridworld MNIST environment, where instead of a $2\times 2$ square there are two MNIST digits \nocite{lecun1998mnist}. The two digits can be moved on the grid via 4 directional actions (so there are 8 actions total), the first digit is always odd and the second digit always even, so they are distiguishable. In Figure \ref{fig:gridworld-only-sel} we plot each latent feature $f_k$ as a curve, as a function of each ground truth. For example we see that the black feature recovers $+x_1$, the horizontal position of the first digit, or that the purple feature recovers $-y_2$, the vertical position of the second digit.

\subsection{Experiments on MazeBase}

\label{sec:exp-mb}
We use MazeBase~\citep{sukhbaatar2015mazebase} to assess the performance of our continuous embeddings approach on a more complex and well-known environment. MazeBase contains 10 different 2D games in which an agent has to solve a specific task (going to a certain location on the board, activate switches, move a block to a specific place...).  
 We do not aim to solve the game, and only deal with one-step policies.

In this setting, the agent (a red circle) can move in a small environment ($64\times 64$ pixels) and perform the actions \texttt{down, left, right, up}, and, to complexify the disentanglement task, we add the redundant action \texttt{up} as well as the action \texttt{down+left}. The agent can go anywhere except on the orange blocks.

In Figure \ref{fig:dis_space}, we show that the learned representation is such that for each underlying factor of variation, the learned representation clusters $dh$ vectors such that it is possible to decompose the variation between two arbitrary state representations as a sum of small variations along a trajectory (Figure \ref{fig:prediction_recovering}).

\subsubsection{Continuous policy embeddings}
\label{sec:exp-mb-cpe}

We consider the model described in \S \ref{sec:polemb}. 
Our architecure is as follows: the encoder, mapping the raw pixel state to a latent representation, is a 4-layer convolutional neural network with batch normalization \citep{ioffe2015batch} and leaky ReLU activations. The decoder uses the transposed architecture with ReLU activations. The noise $z$ is sampled from a 6-dimensional gaussian distribution and both the generator $\Phi(h,z)$ and the policy $\pi(h,\phi)$ are neural networks consisting of 2 fully-connected layers. Our attribute selector $A(dh, \phi)$ is a gaussian kernel. In practice, a minibatch of $n = 64$ vectors $\phi_1, \dots, \phi_{64}$ is sampled at each step. The agent randomly choses one $\phi = \phi_{behavior}$ and samples an action $a \sim \pi(h, \phi_{behavior})$. Our model parameters are then updated using policy gradient and importance sampling. For each selectivity reward, the term $\mathbb{E}_{\phi'}[ | A(h' - h,\phi')|]$ is estimated as $\tfrac{1}{n} \sum_{i = 1}^n | A(h' - h,\phi_i)|$.

\begin{figure}[H]
\centering
\subfloat[]{\includegraphics[width=.4\linewidth]{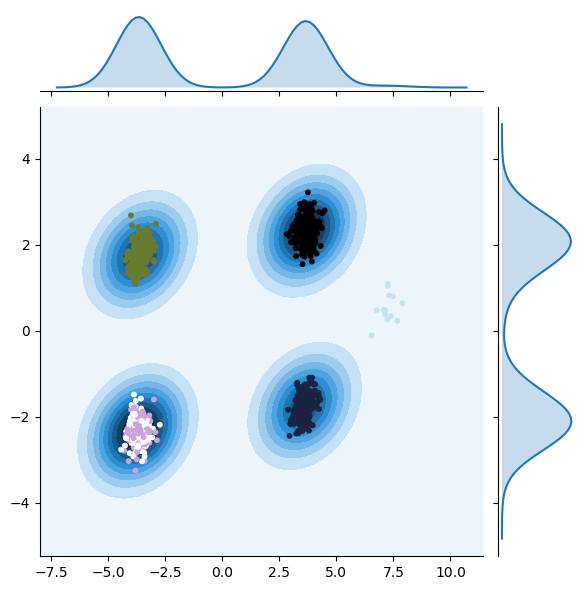}}
\hfill
\subfloat[]{\includegraphics[width=.5\linewidth]{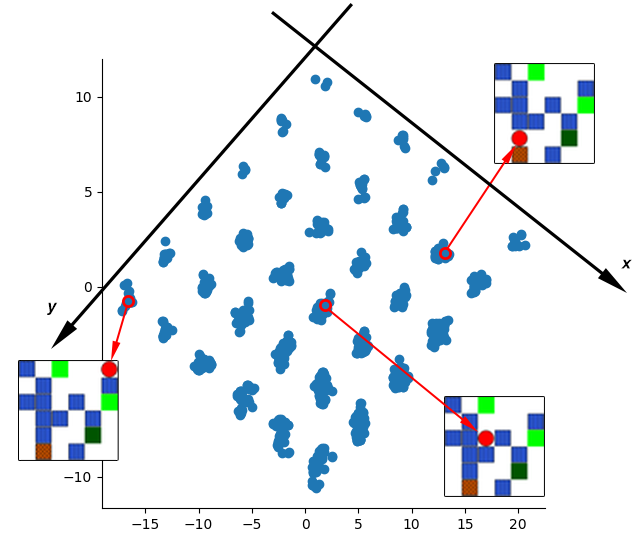}} 

\caption[test]{(a) Sampling of $1000$ variations $dh = h' - h$ and its kernel density estimation encountered when sampling random controllable factors $\phi$. We observe that our algorithm disentangles these representations on $4$ main modes, each corresponding to the action that was actually taken by the agent.\protect\footnotemark
\, (b) The disentangled stucture in the latent space. The $x$ and $y$ axis are disentangled such that we can recover the $x$ and $y$ position of the agent in any observation $s$ simply by looking at its latent encoding $h = f(s)$. The missing point on this grid is the only position the agent cannot reach as it lies on an orange block.}
\label{fig:dis_space}
\end{figure}

After jointly training the reconstruction and selectivity losses, our algorithm disentangles four directed factors of variations as seen in Figure~\ref{fig:dis_space}: $\pm x$-position and $\pm y$-position of the agent. For visualization purposes, in the rest of the section, we chose the bottleneck of the autoencoder to be of size $K = 2$.

The disentanglement appears clearly as the latent features corresponding to the $x$ and $y$ position are orthogonal in the latent space. Moreover, we notice that our algorithm assigns both actions \texttt{up} (white and pink dots in Figure~\ref{fig:dis_space}.a) to the same feature. It also does not create a signifant mode for the feature corresponding to the action \texttt{down+left} (light blue dots in Figure~\ref{fig:dis_space}.a) as this feature is already explained by features \texttt{down} and \texttt{left}.
\footnotetext{pink and white for \texttt{up}, light blue for \texttt{down+left}, green for \texttt{right}, purple black \texttt{down} and night blue for \texttt{left}.}

\subsubsection{Towards planning and policy inference}
\label{sec:exp-mb-ppi}

\begin{figure}[H]
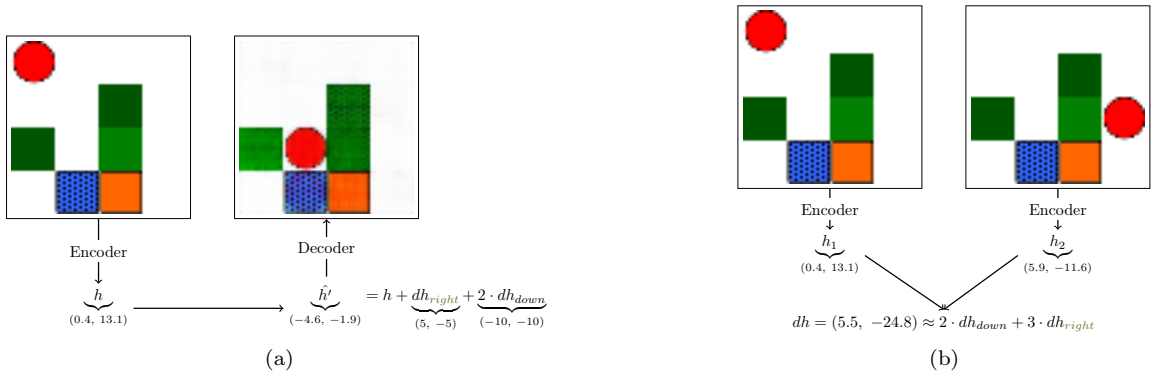

\centering
\subfloat[]{\input{prediction.tex}}
\hfill
\subfloat[]{\input{recovering.tex}} 
\caption{(a) Predicting the effect of a cause on Mazebase. The leftmost image is the visual input of the environment, where the agent is the round circle, and the switch states are represented by shades of green. After the training, we are able to distinguish one cluster per $dh$ (Figure \ref{fig:dis_space}), that is to say per variation obtained after performing an action, independently from the position $h$. Therefore, we are able to move the agent just by adding the corresponding $dh$ to our latent representation $h$. The second image is just the reconstruction obtained by feeding the resulting $h'$ into the decoder. (b) Given a starting state and a goal state, we are able to decompose the difference of the two representations $dh$ into a (non-directed) sequence of movements.}
\label{fig:prediction_recovering}
\end{figure}

This disentangled structure could be used to address many challenging issues in reinforcement learning. We give two examples in figure~\ref{fig:prediction_recovering}: 
\begin{itemize}
\item Model-based predictions: Given an initial state, $s_0$, and an action sequence $a_{\{0:T-1\}}$, 
we want to predict the resulting state $s_T$.

\item A simplified deterministic policy inference problem: Given an initial state $s_{start}$ and a terminal state $s_{goal}$, we aim to find a suitable action sequence $a_{\{0:T-1\}}$ such that $s_{goal}$ can be reached from $s_{start}$ by following it.
\end{itemize}
Because of the $tanh$ activation on the last layer of $\phi(h, z)$, the different factors of variation $dh = h' - h$ are placed on the vertices of a hypercube of dimension $K$, and we can think of the the policy inference problem as finding a path in that simpler space, where the starting point is $h_{start}$ and the goal is $h_{goal}$. We believe this could prove to be a much easier problem to solve.

However, this disentangled representation alone cannot solve completely these two issues in an arbitrary environment. Indeed, the only factors we are able to disentangle are the factors directly \textit{controllable} by the agent, thus, we are not able to account for the ambiant dynamics or other agents' influence.

\section{Related work}
\label{sec:related}
There is a large body of work on learning features in RL focusing on indirectly learning good internal representations. In \cite{jaderberg2016reinforcement}, agents learn off-policy to control their pixel inputs, forcing them to learn features that 
help  control the environment (at the pixel level). \cite{oh2015action} propose models that learn to predict the future, conditioned on action sequences, which push the agent 
to capture temporal features. Many more works go in this direction, such as (deep) successor feature representations \citep{dayan1993improving,kulkarni2016deep} or the options framework \citep{sutton1999between,precup2000temporal} when used in conjunction with neural networks \citep{bacon2016option}.

Our approach is similar in spirit to the Horde architecture~\citep{sutton2009}. In that scenario, agents learn policies that maximize specific inputs, whereas we learn policies that control simultaneously learned
features of the input. The predictions for all these policies then become features for the agent. Our objective is defined specifically in the context of autoencoders but can be generalized to other
representation-learning frameworks. Unlike recent work on the predictron~\citep{silver2017}, our approach is not focused on solving a planning task, and the goal is simply to learn how agents control their environment. 

\section{Conclusion and discussion: Scaling to general environments, controllability and the binding problem}
\label{sec:general}

We have introduced a novel type of clue aiming at learning representations which disentangle the underlying factors of variation. The main assumption is that some of those factors correspond to independently controllable aspects of the environment. This leads to training frameworks in which one learns jointly a set of exploratory policies and corresponding features of the learned representation which disentangle those controlled aspects. This is only a first step towards training agents which learn to control their environments at the same time as learning good representations of it. 

We focused on the simpler setups in which the environment is made of a static set of objects. In this case,  if the objective posited in \S \ref{sec:pol-sel} is learned correctly, we can assume that feature $k$ of the representation can unambiguously refer to some controllable property of some specific object in the environment. For example, the agent's world might contain only a red circle 
and a green rectangle, which are only affected by the actions of the agent (they do not move on their own) and we only change the positions and colours of these objects from one trial 
to the next. Hence, a specific feature $f_k$ can learn to unambiguously refer to the position or the colour of one of these two objects.

In reality, environments are stochastic, and the set of objects  in a given scene is drawn from some distribution. The number of objects may vary and their types may be different. 
It then becomes less obvious how feature $k$ could refer in a clear way to some feature of one of the objects in a particular scene. 
If  we have \textit{instances} of objects of different types, some addressing or naming scheme is required to refer to the particular objects (instances) present in the scene, so as to match the policy with a particular attribute of a particular object to selectively modify. While our proposed distributed alternative (\S \ref{sec:polemb}) is an attempt to address this, a fundamental representational problem remains. 

This is connected to the binding problem in neuro-cognitive science: how to represent a set of objects, each  having different attributes, so that we do not confuse, for example, the set $\{$red circle, blue square$\}$ with $\{$red square, blue circle$\}$. 
The binding problem has received some attention in the representation learning literature \citep{minin2012complex,greff2016tagger}, 
 but still remains mostly unsolved. 
Jointly considering this problem and learning controllable features may prove fruitful.

These ideas may also lead to interesting ways of performing exploration. The RL exploration process could be driven by a notion of controllability, predicting the interestingness of objects in a scene and choosing features and associated policies with which to attempt controlling them -- such ideas have only been briefly explored in the literature (e.g.~\cite{ratitch}).
How do humans choose with which object to play? We are attracted to objects for which we do not yet know if and how we can control them, and such a process may be critical to learn how the world works.



\end{document}